\pdfoutput=1%For ARXIV
\documentclass[letterpaper]{article} % DO NOT CHANGE THIS
\usepackage{aaai24}  % DO NOT CHANGE THIS
\usepackage{times}  % DO NOT CHANGE THIS
\usepackage{helvet}  % DO NOT CHANGE THIS
\usepackage{courier}  % DO NOT CHANGE THIS
\usepackage[hyphens]{url}  % DO NOT CHANGE THIS
\usepackage{graphicx} % DO NOT CHANGE THIS
\usepackage{natbib}  % DO NOT CHANGE THIS AND DO NOT ADD ANY OPTIONS TO IT
\usepackage{caption} % DO NOT CHANGE THIS AND DO NOT ADD ANY OPTIONS TO IT
\usepackage{algorithm}
\usepackage{algorithmic}

\usepackage{newfloat}
\usepackage{listings}
\DeclareCaptionStyle{ruled}{labelfont=normalfont,labelsep=colon,strut=off} % DO NOT CHANGE THIS
\floatstyle{ruled}
\newfloat{listing}{tb}{lst}{}
\floatname{listing}{Listing}
\usepackage[utf8]{inputenc} % allow utf-8 input
\usepackage{url}            % simple URL typesetting
\usepackage{booktabs}       % professional-quality tables
\usepackage{amsfonts}       % blackboard math symbols
\usepackage{nicefrac}       % compact symbols for 1/2, etc.
\usepackage{microtype}      % microtypography
\usepackage{xcolor}         % colors

\usepackage{multirow}
\usepackage{multicol}
\usepackage{diagbox}
\usepackage{amsmath}
\usepackage{graphicx}

\usepackage{caption}
\usepackage{subfigure}
\usepackage{amssymb}
\usepackage{colortbl}
\usepackage{nicematrix}
\newtheorem{definition}{Definition}

\newcommand{\labeledset}{\mathcal{D}_{l}}
\newcommand{\unlabeledset}{\mathcal{D}_{u}}
\newcommand{\testset}[1]{$\mathcal{D}_{\text{\uppercase\expandafter{\romannumeral#1}}}$}
\newcommand{\aVec}[1]{\mathbf{#1}}
\newcommand{\roma}[1]{\uppercase\expandafter{\romannumeral#1}}
\newcommand{\intraset}{\mathcal{D}_{\text{intra}}}
\newcommand{\interset}{\mathcal{D}_{\text{inter}}}

\DeclareMathOperator{\betadist}{Beta}
\DeclareMathOperator*{\expectation}{\mathbb{E}}

\title{NodeMixup: Tackling Under-Reaching for Graph Neural Networks}
\author{
    Weigang Lu, 
    Ziyu Guan,   
    Wei Zhao\thanks{Corresponding author.},
    Yaming Yang,
    Long Jin
}
\affiliations{
    School of Computer Science and Technology, Xidian University, China\\
    \{wglu@stu., zyguan@, ywzhao@mail., yym@, jin@stu.\}xidian.edu.cn
    
}

\usepackage{bibentry}
\begin{document}

\maketitle

\begin{abstract}
Graph Neural Networks (GNNs) have become mainstream methods for solving the semi-supervised node classification problem. However, due to the uneven location distribution of labeled nodes in the graph, labeled nodes are only accessible to a small portion of unlabeled nodes, leading to the \emph{under-reaching} issue. In this study, we firstly reveal under-reaching by conducting an empirical investigation on various well-known graphs. Then, we demonstrate that under-reaching results in unsatisfactory distribution alignment between labeled and unlabeled nodes through systematic experimental analysis, significantly degrading GNNs' performance. To tackle under-reaching for GNNs, we propose an architecture-agnostic method dubbed NodeMixup. The fundamental idea is to (1) increase the reachability of labeled nodes by labeled-unlabeled pairs mixup, (2) leverage graph structures via fusing the neighbor connections of intra-class node pairs to improve performance gains of mixup, and (3) use neighbor label distribution similarity incorporating node degrees to determine sampling weights for node mixup. Extensive experiments demonstrate the efficacy of NodeMixup in assisting GNNs in handling under-reaching. The source code is available at \url{https://github.com/WeigangLu/NodeMixup}.

\end{abstract}

\section{Introduction}
% GNNs
Graph Neural Networks (GNNs)~\cite{gcn, gat, sgc, gcnii, appnp, sage}, which are designed based on the message-passing protocol~\cite{mpnn}, have become the mainstream models for dealing with the semi-supervised node classification problem. A recent work~\cite{distribution} reveals the success of GNNs is that the propagation on graphs narrows the distribution gap between labeled and unlabeled data (distribution alignment), thereby benefiting GNNs to make reasonable inferences over unlabeled data. At the training stage, the model is optimized by minimizing the supervised loss function which is defined on labeled nodes. Then, at the inference stage, the well-trained model makes predictions on unlabeled nodes. 
\begin{figure}[!htbp]
        \centering
       \includegraphics[width=0.85\columnwidth]{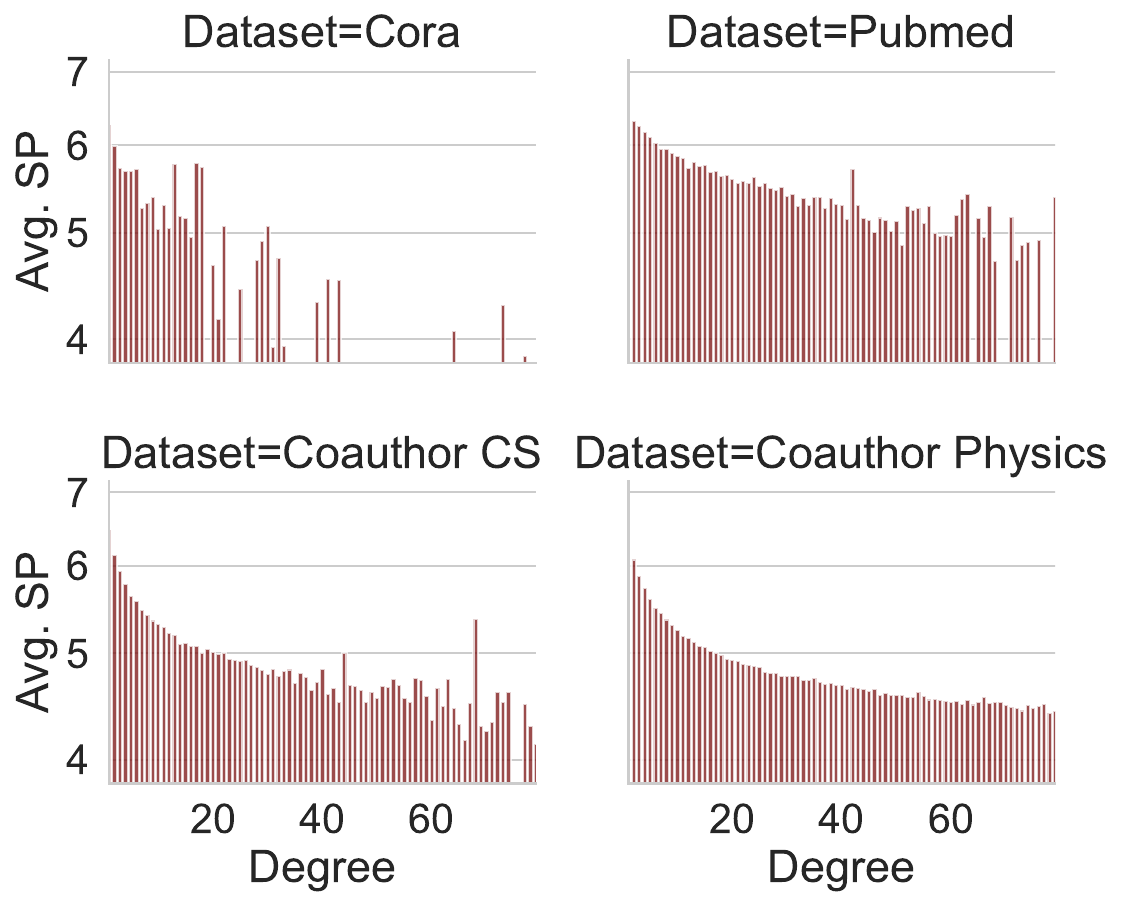}
        \caption{Visual illustrations of under-reaching. Avg. SP is the mean value of the average shortest path length from each unlabeled node to all the labeled ones, where the mean is taken over unlabeled nodes with the same degree. Lower-degree nodes are farther away from labeled nodes while higher-degree nodes tend to be closer to labeled nodes.}
        \label{fig:under-reaching}
\end{figure}
In other words, a $K$-layer GNN helps labeled nodes to receive information from $k$-hop neighbors ($1 \leq k \leq K$) and learns from these labeled nodes to capture a better picture of unlabeled data.

% Introduction to under-reaching
However, by revisiting several commonly-used graphs, we find that nodes with lower (higher) degrees usually stay farther away from (nearer to) labeled nodes, as illustrated in Figure~\ref{fig:under-reaching}. With the restriction of model depth, those lower-degree nodes could hardly transmit information to far-off labeled nodes. Thus, massive unlabeled nodes are hardly known by labeled nodes in popular 2-layer GNN architectures. It leads to incomplete knowledge about unlabeled nodes, hindering distribution alignment. As a result, the trained GNN can only recognize the nodes located near labeled nodes, whereas other unseen-during-training nodes are difficult to be classified. However, in practical scenarios, labeled nodes tend to distribute sparsely in the graph. This phenomenon, namely \emph{under-reaching}~\cite{pastel, over, rewiring, under_reaching}, will be discussed in detail.

There have been several methods attempting to alleviate under-reaching. Intuitively, stacking more GNN layers to allow all nodes to propagate more distant information seems to be a direct solution. Nevertheless, it still raises two additional problems, i.e., over-smoothing~\cite{li2018deeper} that induces indistinguishable node representations and over-squashing~\cite{bottleneck} that causes distant information loss. To increase reachability through modifying the graph structure, \cite{pastel, rewiring} leverage $k$-hop positional encodings to add edges between nodes. Unfortunately, they require substantial computational costs in calculating the shortest path between each node pair. Considering practical use, how to develop an effective and flexible method to increase reachability is still a challenging problem.

% Introduction to Mixup which can alleviate under-reaching
The key insight to tackle under-reaching is to \emph{improve communications between labeled and unlabeled nodes, facilitating distribution alignment in training}. Recently, mixup~\cite{mixup,graphmixup,graphmix,mixup_for_node} techniques have been widely adopted to synthesize additional labeled data via random interpolation between pairs of data points and corresponding labels from original labeled data. The synthesized data can be used as the augmented input for the backbone model. Interestingly, interpolation is similar to the message-passing mechanism since they both essentially perform weighted sum. An intuitive idea is to mix up labeled and unlabeled nodes to enhance their communications. However, the traditional mixup techniques are proposed to expand labeled data but less adept at addressing the under-reaching issue due to the following reasons: (1) The mixed pairs are only sampled from the labeled set which leads to limited access to unlabeled nodes; (2) Traditional mixup methods often employ linear interpolation on data features and labels, which proves less adaptable to the intricate graph topology capturing relationships between nodes in graph-structured data.%the traditional mixup does not take advantage of graph structures where intra-class node pairs tend to share similar neighbor distribution, facilitating intra-class cohesion~\cite{mad}.
 
Inspired by these insights, we develop NodeMixup, an architecture-agnostic method to tackle under-reaching for GNNs. To improve communications between labeled and unlabeled nodes, we propose cross-set pairing that chooses mixed pairs from labeled and pseudo-labeled unlabeled sets~\cite{pl}. Besides, we enhance intra-class interactions by merging neighbor connections among intra-class nodes, and use the standard mixup operation for inter-class node pairs which contributes to characterizing more generalizable classifying boundaries~\cite{mixup}. Notably, we propose a novel Neighbor Label Distribution (NLD)-Aware Sampling, which leverages the similarity of neighborhood label distributions along with node degrees to compute sampling weights. It ensures unlabeled nodes with dissimilar/similar neighbor patterns to labeled nodes, as well as nodes with lower degrees, are more likely to be selected for inter/intra-class mixup. NodeMixup enables direct interactions between node pairs and escapes from the restriction of the graph structure or model depth. This simple but effective approach can be applied to any GNN without complex architecture adjustments or significant computational costs.

\paragraph{Contributions.}
Our main contributions are as follows: (1) We revisit and analyze the under-reaching issue through empirical observations, highlighting its negative impacts on communications between labeled and unlabeled nodes, which hampers distribution alignment. (2) We propose NodeMixup, an architecture-agnostic framework that facilitates direct communications between labeled and unlabeled nodes, overcoming the limitations imposed by the graph structure and effectively alleviating under-reaching for GNNs. (3) We apply NodeMixup on popular GNNs and evaluate it on six real-world graphs. It consistently achieves significant performance gains for GNNs, showing its practicality and generalizability. 

\section{Preliminary and Related Works}
\label{sec:related}
\subsection{Notations}
We use $\mathcal{G}=\{\mathcal{V}, \mathcal{E}, A\}$ to denote an undirected graph with self-loops, where $\mathcal{V}=\{\aVec{x}_{1}, \cdots, \aVec{x}_{N}\}$ is the node set with $N$ nodes, $\mathcal{E}$ is the edge set and $A \in \mathbb{R}^{N \times N}$ is the adjacency matrix. We also have the input feature matrix $X \in \mathbb{R}^{N \times F}$ whose $i$-th row vector is denoted as $\aVec{x}_{i}$, where $F$ is the input dimensionality. Here, we abuse the notation $\aVec{x}_{i}$ a bit to denote it as the node index of node $i$. We define the node label matrix as $Y \in \mathbb{R}^{N \times C}$, where $C$ is the number of classes and $\aVec{y}_{i}$ is the one-hot encoding of the label of node $i$. We divide the data set into labeled set $\labeledset=\{(\aVec{x}_{1},\aVec{y}_{1}) \cdots, (\aVec{x}_{N_{l}}, \aVec{y}_{N_{l}})\}$ and unlabeled set $\unlabeledset=\{\aVec{x}_{N_{l}+1}, \cdots, \aVec{x}_{N}\}$, where $0 < N_{l} < N$. Specifically, the training set can be divided into $C$ subsets according to different classes, i.e., $\mathcal{V}_{l}^{(1)}, \cdots, \mathcal{V}_{l}^{(C)}$. Besides, we assume that each subset contains $T$ labeled nodes.

\subsection{Related Works}
\subsubsection{Graph Neural Networks.}
\cite{sage,gcn,gat,gin,sgc,gcnii,appnp,gprgnn} enable each node to accept the information from neighbors in the range of $K$ hops. The discernible difference is how they aggregate messages from connected nodes at each layer. Assuming $\aVec{h}^{(l)} \in \mathbb{R}^{F}$ is a $F$-dimension representation, the $l$-th GNN layer $f^{(l)}$ aggregates and transforms neighbor information to produce $\aVec{h}^{(l+1)} \in \mathbb{R}^{F}$, which can be generalized as:
$
	\aVec{h}^{(l+1)} = f^{(l)}(\aVec{h}^{(l)}, \theta^{(l)}, A),
$
where $\theta^{(l)} \in \mathbb{R}^{F \times F} $ is the learnable parameter and $\sigma$ is an activation function. Without loss of generalizability, we can define a $L$ layer GNN as $G_{\theta}(\aVec{x}, A) = (f^{(L)} \circ f^{(L-1)} \circ \dots \circ f^{(1)})_{\theta} (\aVec{x}, A)$, which is parameterized by $\theta$. Therefore, the cross-entropy loss~\cite{cross_entropy} $\ell$ can be adopt in semi-supervised node classification as: 
\begin{small}
	\begin{equation}
	\label{loss-gnn}
		\mathcal{L}_{\text{GNN}} (\labeledset, G_{\theta}, A) = \expectation_{(\aVec{x}, \aVec{y}) \sim \labeledset} \ell(G_{\theta}(\aVec{x}, A), \aVec{y}).
	\end{equation} 
\end{small}

\subsubsection{Mixup.}
\cite{mixup} proposes the mixup technique that mixes features and corresponding labels of pairs of labeled samples to generate virtual training data. Because of the simplicity and effectiveness of mixup, some works~\cite{mixup_metric_g,mixup_transplant_g,ifmixup_g,graphmad_g} adapt it to the graph domain. However, they only focus on the graph classification problem and can not be directly applied to the node-level task. 
\begin{figure*}[!htbp]
	\centering
    \includegraphics[width=1\linewidth]{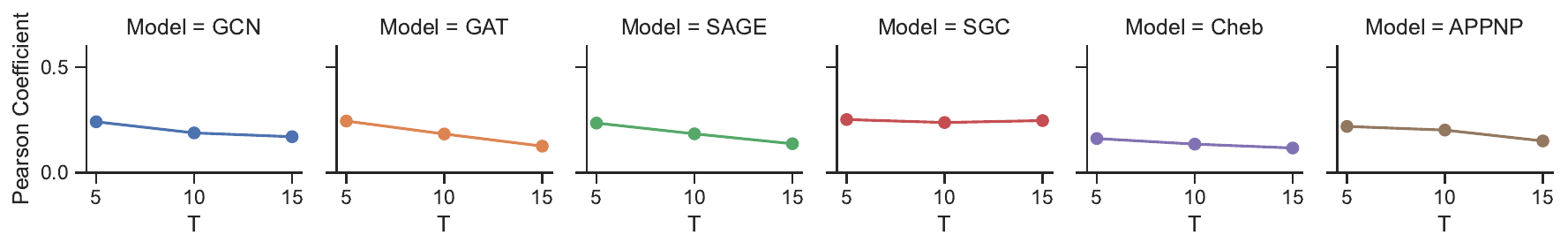}
    \caption{The correlation between prediction scores on actual classes and RC values on \textsc{Cora} dataset. $T = m$ represents $m$ labeled nodes per class. The Pearson coefficient shows a positive correlation between prediction scores and RC, demonstrating that larger reachability yields better performance. As labeled nodes decrease which (indirectly) suggests poor reachability, a more significant positive correlation can be observed.}
    \label{fig:pearson}
\end{figure*}
To overcome the node classification problem, \cite{graphmixup,mixup_for_node,graphmix} develop improved mixup mechanisms to enhance GNNs. Formally, assuming both $\aVec{a}$ and $\aVec{b}$ are either feature vectors or one-hot label vectors, the mixup operation is defined as:
\begin{small}
	\begin{equation}
	\label{eq:mixup}
	\mathcal{M}_{\lambda}(\aVec{a}, \aVec{b}) = \lambda \aVec{a} + (1-\lambda) \aVec{b},
	\end{equation}
\end{small}where $\lambda$ is sampled from $\betadist(\alpha, \alpha)$ distribution and $\alpha \geq 0$. 

\subsubsection{Under-reaching, Over-smoothing, and Over-squashing.}
They are the graph-specific information shortage issues in the context of semi-supervised node classification. From a topological perspective, prior researchs have described the \underline{under-reaching} issue~\cite{pastel,rewiring,under_reaching} as a node's inability to be aware of nodes that are farther away than the number of layers. However, directly increasing the number of layers gives rise to the \underline{over-smoothing} issue~\cite{li2018deeper,oono2020graph,skipnode}, where node representations become indistinguishable, severely impacting prediction performance. Even with the resolution of over-smoothing by enlarging receptive fields, the \underline{over-squashing} issue~\cite{bottleneck,over,understanding-os} still persists. This issue pertains to the loss of information from distant nodes due to message propagation over the graph-structured data, where features from exponentially-growing neighborhoods are compressed into fixed-length node vectors. Drawing inspiration from the distribution shift concept~\cite{distribution_shift}, where the difference between labeled and unlabeled distributions affects the model's generalization, we identify the under-reaching issue as a lack of communication between labeled and unlabeled nodes, leading to difficulties in making accurate inferences over unlabeled nodes. Thus, this issue represents a more generalized graph-specific challenge about how to improve communication between labeled and unlabeled nodes rather than propagate distant information in the semi-supervised node classification regime.

\section{Method}
In this section, we first explain our motivation by introducing under-reaching. Then, we present our proposed NodeMixup framework in order to increase the reachability of GNNs.

\subsection{Motivation: Understanding Under-reaching}
\label{sec:under-reaching}

\subsubsection{How Does Under-reaching Impact on GNNs?}
Nodes far from labeled nodes lack supervision information because the influence of labeled nodes decreases with topology distance~\cite{influence_decay}. With the restriction of model depth, nodes at $r$-hop away ($r > K$) from labeled nodes can not be reached when a $K$-layer model (e.g., GCN) is used. Since the supervised loss function is purely defined on labeled nodes, the optimization might be misled by the inadequate received information. We define $ d_{\mathcal{G}}(i,j)$ as the shortest path length between node $i$ and node $j$. To measure reachability for each node, we first introduce the reaching coefficient (\textbf{RC}) from~\cite{pastel} as:
\begin{small}
\begin{equation}
\label{eq:rc}
	\text{RC}_{i} = \frac{1}{|\labeledset|} \sum_{j \in \labeledset} \left( 1 - \frac{\log |d_{\mathcal{G}}(i,j)|}{\log D_{\mathcal{G}}} \right),
\end{equation}	
\end{small}where $D_{\mathcal{G}}$ represents the diameter of graph $\mathcal{G}$, and $d_{\mathcal{G}}(i,j) = D_{\mathcal{G}}$ when node $i$ and node $j$ do not belong to the same connected component. A larger RC value (scaled to $[0,1)$) indicates greater reachability of node $i$. In Figure~\ref{fig:pearson}, we visualize the correlation between prediction scores on actual classes and RC values on \textsc{Cora} dataset using different GNNs, i.e., \textbf{GCN}~\cite{gcn}, \textbf{GAT}~\cite{gat}, \textbf{APPNP}~\cite{appnp}, \textbf{Cheb}Net~\cite{chebnet}, and Graph\textbf{SAGE}~\cite{sage}. Across all experiments, we vary the number of training nodes per class ($T \in {5, 10, 15}$). We can observe positive correlations (Pearson Coefficient larger than 0) in all the cases, which further demonstrates the benefit of better reachability. Additionally, as $T$ decreases, indicating that the reachability declines, the positive correlation becomes more significant. It is because only a few unlabeled nodes can be seen during training. It is easier for GNNs to classify correctly those unlabeled nodes located nearby labeled nodes.
 
\begin{figure*}[!ht]
	\centering
    \includegraphics[width=0.95\linewidth]{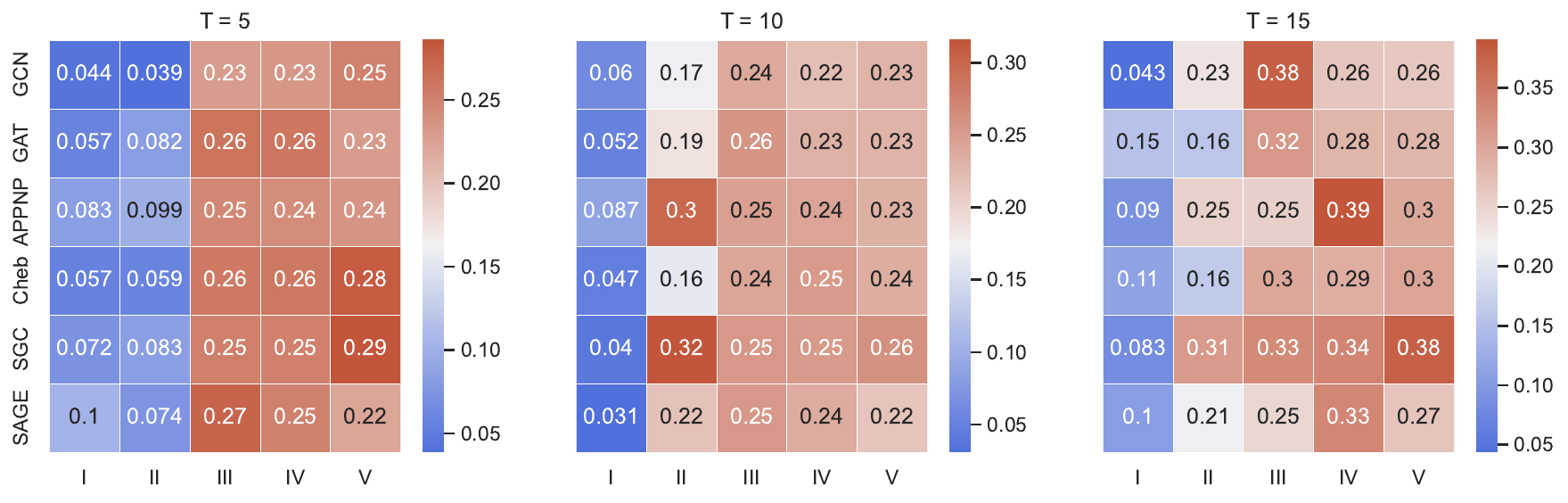}
    \caption{The heat map of CKA between labeled and unlabeled distributions learned by different GNNs using \textsc{Cora}. A larger CKA value indicates more similar representation distributions between labeled and unlabeled nodes. \roma{1}, ..., \roma{5} represent unlabeled nodes with various reachability arranged in ascending order. Larger reachability is in favor of narrowing the distribution gap.}
    \label{fig:cka}
\end{figure*}

 \subsubsection{Why Does Under-reaching Fail GNNs?} 
 A recent study~\cite{distribution} underscores that GNN success hinges on aligning the distributions of labeled and unlabeled nodes. The propagation enables labeled nodes to receive information from unlabeled nodes to narrow the distribution gap between labeled and unlabeled nodes. Intuitively, it would facilitate inference as the two distributions get close. Inspired by this, to further understand the negative impact of under-reaching, we here investigate the distance between the labeled and unlabeled distribution at different levels of RC. We first briefly introduce the centered kernel alignment (CKA)~\cite{cka} metric\footnote{Please refer to~\cite{cka, distribution} for more details about CKA.}, which is widely used to measure the representation similarity. Supposing $Z_{l}, Z_{u} \in \mathbb{R}^{m \times n}$ are the representations (learned by an arbitrary GNN) sampled from labeled and unlabeled nodes, the distribution similarity between $Z_{l}$ and $Z_{u}$ is measured by CKA as:
 \begin{small}
 	 \begin{equation}
 	\text{CKA}(Z_{l}, Z_{u}) = \frac{\lVert Z_{u}^{T}Z_{l}\rVert_{F}}{\lVert Z_{l}Z_{l}^{T} \rVert_{F} \lVert Z_{u}Z_{u}^{T} \rVert_{F}}.
 \end{equation}
 \end{small}A larger CKA (scaled to $[0,1]$) implies a higher similarity. Secondly, we divide unlabeled nodes from \textsc{Cora} into five subsets \testset{1}, ..., \testset{5} according to different interval ranges of RC, i.e., range \roma{1}, ..., range \roma{5}\footnote{Let RC$_{max}$ be the maximum value of RC in all the unlabeled nodes. Then, we define range \roma{1} as $[0, \nicefrac{\text{RC}_{max}}{5}]$, range \roma{2} as $(\nicefrac{\text{RC}_{max}}{5}, \nicefrac{2\text{RC}_{max}}{5}]$, ..., and range \roma{5} as $(\nicefrac{4\text{RC}_{max}}{5}, \text{RC}_{max}]$.}. Finally, we calculate CKA between $Z_{l}$ and representations of unlabeled nodes from \testset{1}, ..., \testset{5} learned by different GNNs. To do so, we sample the same number of nodes in each pair of two sets ($\labeledset$, \testset{1}), ..., ($\labeledset$, \testset{5}), and the number is determined by the minimum element number of each set pair. We visualize the results in Figure~\ref{fig:cka}, in which each block represents the CKA value between labeled and unlabeled distributions learned by various GNNs. We can see that lower reachability (e.g., range \roma{1} and \roma{2}) tends to result in a larger distribution gap while higher reachability can bridge the gap.

\subsubsection{How Do We Alleviate Under-reaching?}
From the above analysis, we can see that under-reaching hinders the distribution alignment since the labeled nodes can only reach a small part of unlabeled nodes. A straightforward idea for assisting labeled nodes in reaching more unlabeled nodes is to add edges between them or stack more GNN layers. However, the edge-adding strategy could induce prohibitive computation costs for finding globally friendly neighbors~\cite{rewiring, pastel} or lead to noisy graphs without sufficient supervision~\cite{pastel}. Besides, deepening GNNs leads to over-smoothing or over-squashing. Based on these limitations, we intend to develop an efficient framework for various GNNs to tackle under-reaching. Recently, interpolation-based methods~\cite{mixup, graphmixup,mixup_for_node,graphmix} show great effectiveness and flexibility in augmenting labeled data. Inspired by this, we propose NodeMixup, which mixes labeled and unlabeled data to increase reachability for GNNs. We present the detailed methodology of our NodeMixup in the following section.

\subsection{Methodology}
\label{sec:nodemixup}
 \begin{figure*}[!ht]
 	\centering
    \includegraphics[width=1\linewidth]{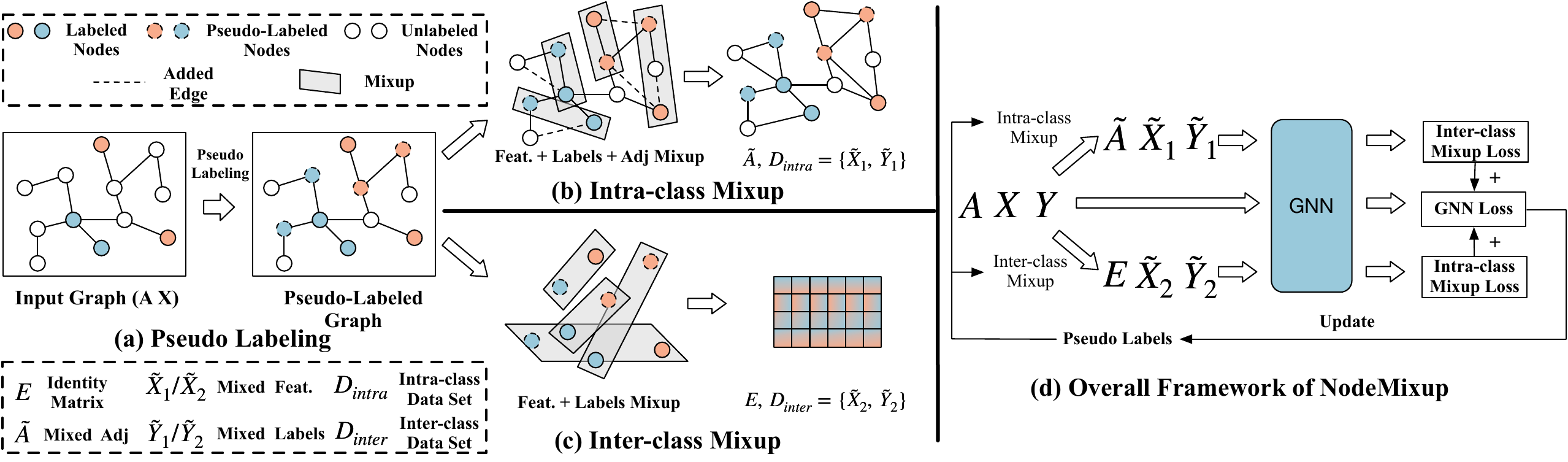}
    \caption{The pipeline of NodeMixup. The NLD-aware sampling is omitted in this figure.}
    \label{fig:nodemixup}
\end{figure*}

We start by giving a high-level overview of NodeMixup and present its pipeline in Figure~\ref{fig:nodemixup}. Our NodeMixup is designed as a data augmentation technique that can be easily applied to various GNNs. At each iteration, NodeMixup samples node pairs from labeled and unlabeled nodes. This cross-set pairing ensures labeled nodes reach more unlabeled nodes to bridge the distribution gap. Then, we use the pseudo labels~\cite{pl} with a confidence threshold $\gamma$ for unlabeled nodes and construct a pseudo-labeled data set $\mathcal{D}_{pl}$. Guided by pseudo labels, we mix nodes via different mixup operations, i.e., intra-class and inter-class mixup. Then, the intra-class-mixup nodes with the mixed adjacency matrix $\tilde{A}$ and inter-class-mixup nodes with an identity matrix $E \in \mathbb{R}^{N_{l} \times N_{l}}$ are used as inputs for the GNN model, where $N_{l}$ is the number of labeled nodes. Note that a GNN fed with an identity matrix equals a MLP. Besides, to enhance the mixup effect, we propose a neighbor label distribution (NLD)-aware sampling strategy. It ensures that nodes with similar neighbor patterns and lower degrees are more likely to be selected. Finally, the model is optimized via minimizing the loss $\mathcal{L}$ from the combination of $\mathcal{L}_{\text{GNN}}$, $\mathcal{L}_{\text{intra-M}}$, and $\mathcal{L}_{\text{inter-M}}$ (defined by Eq.~(\ref{loss-gnn}), Eq.~(\ref{loss-intra-mixup}), and Eq.~(\ref{loss-inter-mixup}), respectively) from each branch as follows:
\begin{small}
	\begin{equation}
	\mathcal{L} = \mathcal{L}_{\text{GNN}} + \lambda_{\text{intra}}\mathcal{L}_{\text{intra-M}} + \lambda_{\text{inter}}\mathcal{L}_{\text{inter-M}},
\end{equation}
\end{small}where two hyper-parameters $\lambda_{\text{intra}}, \lambda_{\text{inter}} \in (0,1]$ control the regularization effect of NodeMixup.

\subsubsection{Intra-class Mixup.}
 We sample and mix nodes with the equal class from $\labeledset$ and $\mathcal{D}_{pl}$ via Eq.~(\ref{eq:mixup}). Suppose $\aVec{x}_{i} \in \labeledset$ and $\aVec{x}_{j} \in \mathcal{D}_{pl}$ are classified as $\aVec{y}_{i}$ and $\hat{\aVec{y}}_{j}$, where $\hat{\aVec{y}}_{j}$ is the pseudo label of node $j$. We have the mixed intra-class data set $\intraset$ as:
 \begin{small}
 	\begin{equation}
 	\label{D-intra}
 	\begin{split}
 	\intraset = &\{(\tilde{\aVec{x}}_{i}, \tilde{\aVec{y}}_{i} )|\tilde{\aVec{x}}_{i} = \mathcal{M}_{\lambda}(\aVec{x}_{i}, \aVec{x}_{j}),  \tilde{\aVec{y}}_{i} = \mathcal{M}_{\lambda}(\aVec{y}_{i}, \hat{\aVec{y}}_{j}), \\
 				&(\aVec{x}_{i}, \aVec{y}_{i}) \in \labeledset, (\aVec{x}_{j}, \aVec{y}_{j}) \in \mathcal{D}_{pl}, \aVec{y}_{i} = \hat{\aVec{y}}_{j}\}.
 	\end{split}
 	\end{equation}
 \end{small}
Furthermore, we also fuse their topological information together since these intra-class pairs tend to share similar neighbor distribution, which prompts intra-class similarity. Thus, we mix the topological information of node $i$ and node $j$ to generate a new adjacency matrix $\tilde{A}$ as: 
\begin{small}
	\begin{equation}
	\begin{split}
		&\text{Row mixup:}\quad\tilde{A}_{i, :} = \mathcal{M}_{\lambda}(A_{i, :}, A_{j, :}),\\
		&\text{Column mixup:}\quad\tilde{A}_{:, i} = \mathcal{M}_{\lambda}(A_{:, i}, A_{:, j}).
	\end{split}
	\end{equation}
\end{small}
Here, $A_{i, :}$ and $A_{:, i}$ represents the $i$-th row and column vector of $A$. Finally, we calculate the intra-class mixup loss $\mathcal{L}_{\text{intra-M}}$ with a GNN model $G_{\theta}$ as follows:
\begin{small}
	\begin{equation}
	\label{loss-intra-mixup}
		\mathcal{L}_{\text{intra-M}} = \mathcal{L}_{\text{GNN}} (\intraset, G_{\theta}, \tilde{A}) = \expectation_{(\tilde{\aVec{x}}, \tilde{\aVec{y}}) \sim \intraset} \ell(G_{\theta}(\tilde{\aVec{x}}, \tilde{A}), \tilde{\aVec{y}}).
	\end{equation}
\end{small}

\subsubsection{Inter-class Mixup.}
The mixup operation facilitates a linear behavior of the model between selected nodes, thus improving its generalization~\cite{mixup}. Besides, empirical observations indicate that interpolating between inputs with different labels results in decision boundaries that transition linearly from one class to another, reducing prediction errors. Thus, the inter-class mixup operation enhances the model's ability to distinguish boundaries between different classes. By combining nodes from different classes, the model can effectively learn shared features and differences between classes, thereby improving its generalization capability. Based on this, we generate the mixed inter-class data set $\interset$ as:
\begin{small}
	\begin{equation}
	\label{D-inter}
	\begin{split}
		 \interset = &\{(\tilde{\aVec{x}}_{i}, \tilde{\aVec{y}}_{i} )|\tilde{\aVec{x}}_{i} = \mathcal{M}_{\lambda}(\aVec{x}_{i}, \aVec{x}_{j}),  \tilde{\aVec{y}}_{i} = \mathcal{M}_{\lambda}(\aVec{y}_{i}, \aVec{\hat{y}}_{j}), \\
		 &(\aVec{x}_{i}, \aVec{y}_{i}) \in \labeledset, (\aVec{x}_{j}, \aVec{y}_{\hat{j}}) \in \mathcal{D}_{pl}, \aVec{y}_{i} \neq \aVec{\hat{y}}_{j}\}.
	\end{split}
	\end{equation}
\end{small}Furthermore, we argue that inter-class interaction could probably lead to noisy learning. Therefore, we feed the backbone GNN with an identity matrix $E$ to block the message passing between inter-class nodes. Similar to Eq.~(\ref{loss-intra-mixup}), we define the inter-class mixup loss $\mathcal{L}_{\text{inter-M}}$ as follows:
\begin{small}
	\begin{equation}
	\label{loss-inter-mixup}
		\mathcal{L}_{\text{inter-M}} = \mathcal{L}_{\text{GNN}} (\interset, G_{\theta}, E) = \expectation_{(\tilde{\aVec{x}}, \tilde{\aVec{y}}) \sim \interset} \ell(G_{\theta}(\tilde{\aVec{x}}, E), \tilde{\aVec{y}}).
	\end{equation}
\end{small}

\subsubsection{NLD-aware Sampling.}
In a graph, intra/inter-class nodes often form cohesive/incohesive communities or clusters, leading to similar/dissimilar neighbor distributions. The characteristics of a node are not only determined by itself but also influenced by its neighbors. To capture such neighborhood information, we adapt Neighborhood Label Distribution (NLD)~\cite{NLD} and define it in the next.
\begin{definition}{(Neighborhood Label Distribution (NLD))}
	 Given $\bar{Y}$ is the label matrix for all the nodes (the label of unlabeled nodes are their predictions), the neighborhood label distribution of node $i$ is defined as $\aVec{q}_{i} = \frac{1}{|\mathcal{N}_{i}|} \sum_{v \in \mathcal{N}_{i}} \bar{Y}_{v,:}$, where $\mathcal{N}_{i}$ is the neighbor set of node $i$.
\end{definition}
Besides, we employ the sharpening~\cite{sharp} trick to enhance the contrast between class probabilities in NLD. Specifically, we calculate the new distribution $\aVec{q}^{\prime}_{ij}$ as following:
\begin{small}
\begin{equation}
	\aVec{q}^{\prime}_{ij} = \frac{\aVec{q}^{\nicefrac{1}{\tau}}_{i}}{\sum_{k=1}^{C} \aVec{q}^{\nicefrac{1}{\tau}}_{ik}},
\end{equation}	
\end{small}where $0 < \tau \leq 1 $ is the temperature parameter that controls the sharpness of the distribution. As $\tau \to 0$, it leads to a sharper probability distribution. Then, we use the cosine similarity $s_{ij}$ between $\aVec{q}_{i}^{\prime}$ and $\aVec{q}_{j}^{\prime}$ to determine the likelihood of node $i$ and node $j$ sharing a similar neighbor pattern, where 
 \begin{small}
	\begin{equation}
	s_{ij} = \frac{\aVec{q^{\prime}}_{i}^{T}\aVec{q^{\prime}}_{j}}{\lVert \aVec{q^{\prime}}_{i} \rVert_{2} \lVert \aVec{q^{\prime}}_{j} \rVert_{2}}\end{equation}
\end{small}Additionally, we take node degrees into account since low-degree nodes suffer more from under-reaching. Therefore, for a labeled node $i$, the sampling weight $w_{ij}$ of an unlabeled node $j$ to be mixed is defined as follows:
\begin{small}
	\begin{equation}
	w_{ij} = 
	\begin{cases}
	\frac{1}{1 + \beta_{d}d_{j}}e^{\beta_{s} s_{ij}}, & \text{if} \quad \aVec{y}_{i} = \hat{\aVec{y}}_{j}, \\
   \frac{1}{1 + \beta_{d}d_{j}}e^{-\beta_{s} s_{ij}}, & \text{if} \quad \aVec{y}_{i} \neq \hat{\aVec{y}}_{j},	
	\end{cases}
\end{equation}
\end{small}where $d_{j}$ is the degree of node $j$, and $\beta_{s} > 0$ and $\beta_{d} > 0$ control the strength of NLD similarity and node degree, respectively. The term $\nicefrac{1}{1 + \beta_{d}d_{j}}$ ensures that the influence of node degree on the sampling weight is monotonic, leading to a higher sampling weight for low-degree nodes and vice versa. This sampling weight calculation balances the effect of node similarity and node degree, resulting in a reasonable and
\begin{table*}[!htbp]
    \centering
    \begin{tabular}{l l c c c c c}
    \hline
    \toprule
    Models & Strategy & \textsc{Cora} & \textsc{Citeseer} & \textsc{Pubmed} & \textsc{CS} & \textsc{Physics}\\
    \midrule
    \multirow{6}{*}{GCN} & 
    	Original & 81.57\small{$\pm$0.4} & 70.50\small{$\pm$0.6} & 77.91\small{$\pm$0.3} & 91.24\small{$\pm$0.4} & 92.56\small{$\pm$1.3} \\
    ~ & UPS & 82.35\small{$\pm$0.4} & 72.82\small{$\pm$0.6} & 78.45\small{$\pm$0.4} & 91.62\small{$\pm$0.3} & 93.01\small{$\pm$0.3} \\
    ~ & PASTEL & 81.97\small{$\pm$0.6} & 71.32\small{$\pm$0.4} & 78.92\small{$\pm$0.2} & 91.76\small{$\pm$0.6} & $>$ 3 days \\
    ~ & ReNode & 81.98\small{$\pm$0.6} & 69.48\small{$\pm$0.4} & 78.13\small{$\pm$0.7} &	91.32\small{$\pm$0.1} & OOM \\
    ~ & GraphMix & 82.29\small{$\pm$3.7} & \textbf{74.55\small{$\pm$0.5}} & \textbf{82.82\small{$\pm$0.5}} & 91.90\small{$\pm$0.2} & 90.43\small{$\pm$1.7} \\
    ~ & \textbf{NodeMixup} & \textbf{83.47\small{$\pm$0.3}} & 74.12\small{$\pm$0.3} & 81.16\small{$\pm$0.2}  & \textbf{92.69\small{$\pm$0.4}} & \textbf{93.97\small{$\pm$0.4}}\\
    \midrule
    \multirow{6}{*}{GAT} & Original & 82.04\small{$\pm$0.6} & 71.82\small{$\pm$0.8} & 78.00\small{$\pm$0.7} & 90.52\small{$\pm$0.4} & 91.97\small{$\pm$0.6}\\
    ~ & UPS & 82.17\small{$\pm$0.5} & 72.97\small{$\pm$0.7} & 78.56\small{$\pm$0.9} & 91.26\small{$\pm$0.4} & 92.45\small{$\pm$1.1} \\
    ~ & PASTEL & 82.21\small{$\pm$0.3} & 72.35\small{$\pm$0.8} & 78.74\small{$\pm$0.9} & 90.31\small{$\pm$0.2} & $>$ 3 days\\
    ~ & ReNode & 81.88\small{$\pm$0.7} & 71.73\small{$\pm$1.2} & 79.68\small{$\pm$0.5} & 88.36\small{$\pm$0.5} & OOM \\
    ~ & GraphMix & 82.76\small{$\pm$0.6} & 73.04\small{$\pm$0.5} &  78.82\small{$\pm$0.4} & 90.57\small{$\pm$1.0} & 92.90\small{$\pm$0.4} \\
    ~ & \textbf{NodeMixup} & \textbf{83.52\small{$\pm$0.3}} & \textbf{74.30\small{$\pm$0.1}} &  \textbf{81.26\small{$\pm$0.3}} &  \textbf{92.69\small{$\pm$0.2}}  &  \textbf{93.87\small{$\pm$0.3}}\\
    \midrule
    \multirow{6}{*}{APPNP} & Original & 80.03\small{$\pm$0.5} & 70.30\small{$\pm$0.6} & 78.67\small{$\pm$0.2} & 91.79\small{$\pm$0.5} & 92.36\small{$\pm$0.8}\\
    ~ & UPS & 81.24\small{$\pm$0.6} & 71.02\small{$\pm$0.7} & 78.69\small{$\pm$0.7} & 91.77\small{$\pm$0.3} & 92.31\small{$\pm$0.5} \\
    ~ & PASTEL & 81.56\small{$\pm$0.3} & 70.68\small{$\pm$0.8} & 78.39\small{$\pm$0.2} & 91.98\small{$\pm$0.4} & $>$ 3 days\\
    ~ & ReNode & 81.12\small{$\pm$0.2} & 70.04\small{$\pm$0.8} & 78.58\small{$\pm$0.3} &	91.99\small{$\pm$0.2} & OOM \\
    ~ & GraphMix & 82.98\small{$\pm$0.4} & 70.26\small{$\pm$0.4} & 78.73\small{$\pm$0.4} & 91.53\small{$\pm$0.6} & 94.12\small{$\pm$0.1} \\
    ~ & \textbf{NodeMixup} &  \textbf{83.54\small{$\pm$0.4}}  & \textbf{75.12\small{$\pm$0.3}}  & \textbf{79.93\small{$\pm$0.1}} &  \textbf{92.82\small{$\pm$0.2}}  & \textbf{94.34\small{$\pm$0.2}}\\
    \midrule
    \multirow{6}{*}{GraphSAGE} & Original & 78.12\small{$\pm$0.3} & 68.09\small{$\pm$0.8} & 77.30\small{$\pm$0.7} & 91.01\small{$\pm$0.9} & 93.09\small{$\pm$0.4}\\
    ~ & UPS & 81.83\small{$\pm$0.3} & 70.29\small{$\pm$0.6} & 77.82\small{$\pm$0.6} & 91.35\small{$\pm$0.4} & 93.20\small{$\pm$0.4} \\
    ~ & PASTEL & 78.58\small{$\pm$0.6} & 70.31\small{$\pm$0.3} & 78.26\small{$\pm$0.7} & 91.77\small{$\pm$0.6} & $>$ 3 days\\
    ~ & ReNode & 76.48\small{$\pm$1.0} & 70.79\small{$\pm$0.9} & 78.67\small{$\pm$1.2} &	89.61\small{$\pm$0.7} & OOM \\
    ~ & GraphMix & 80.09\small{$\pm$0.8} & 70.97\small{$\pm$1.2} & 79.85\small{$\pm$0.4} & 91.55\small{$\pm$0.3} & 93.25\small{$\pm$0.3} \\
    ~ & \textbf{NodeMixup} & \textbf{81.93\small{$\pm$0.2}} &  \textbf{74.12\small{$\pm$0.4}}  & \textbf{79.97\small{$\pm$0.5}}  & \textbf{91.97\small{$\pm$0.2}} & \textbf{94.76\small{$\pm$0.2}}\\
    \bottomrule
    \hline
    \end{tabular}
    \caption{Node Classification Results on Medium-scale Graphs.}
    \label{tab:results_public}
\end{table*}
interpretable mechanism for the Mixup operation between nodes in the graph. 

\subsubsection{Complexity.}
The mixup operation (Eq.~(\ref{eq:mixup})), encompassing both nodes and labels, centers on matrix addition, which incurs negligible computational overhead due to its parallelizability across vectors. The predominant computational expenditure lies within the NLD-aware sampling phase, which unfolds in three sequential steps: NLD calculation, NLD cosine similarity evaluation, and sampling weight computation. The computational cost of NLD computation is determined by the number of edges in the graph, rendering it $\mathcal{O}(|\mathcal{E}|)$. When determining the NLD-similarity between labeled and selected unlabeled nodes, the complexity equates to $\mathcal{O}(|\mathcal{D}_{pl}|^{2})$, where $|\mathcal{D}_{pl}|$ is the subset of confident nodes used for mixup. Subsequently, the computation of sampling weights for these unlabeled nodes involves $\mathcal{O}(|\mathcal{D}_{pl}|)$ operations. Overall, the integration of NodeMixup within GNNs introduces a marginal computational burden $\mathcal{O}(|\mathcal{E}| + |\mathcal{D}_{pl}|^{2} + |\mathcal{D}_{pl}|)$ because $|\mathcal{D}_{pl}|$ is much smaller than $|\mathcal{E}|$, preserving the overall efficiency of the model.

\section{Experiments}
We evaluate NodeMixup on the semi-supervised node classification task. We  use five medium-scale datasets, i.e., \textsc{Cora}, \textsc{Citeseer}, and \textsc{Pubmed}~\cite{cora}, \textsc{Coauthor CS} and \textsc{Coauthor Physics}~\cite{coauthor}, and a large-scale graph, i.e., \textsc{ogbn-arxiv}~\cite{ogb}. Besides, we do detailed ablation experiments to probe into the design of NodeMixup. 

\subsection{Evaluation on Medium-scale Graphs}
\paragraph{Settings.}
 Similar to the experimental setup of~\cite{pastel}, we choose GCN~\cite{gcn}, GAT~\cite{gat}, APPNP~\cite{appnp}, and GraphSAGE~\cite{sage} as backbone models. For the comparing strategies, we choose PASTEL~\cite{pastel}, ReNode~\cite{renode}, and GraphMix~\cite{graphmix}. PASTEL and ReNode address the topology imbalance issue. GraphMix is a data augmentation method that mixes hidden node representations. UPS~\cite{ups} is a state-of-the-art (SOTA) pseudo-labeling method that generates pseudo-labels with uncertainty-aware selection. We simply apply grid search for all the models since we do not intend to achieve new SOTA performance. We keep the parameters of comparing strategies as they are in the original papers or reference implementations. For our proposed NodeMixup, which is implemented by PyTorch Geometric Library~\cite{torch_geo} with the Adam optimizer~\cite{adam}, we search both $\lambda_{\text{inter}}$ and $\lambda_{\text{intra}}$ in $\{1, 1.1, \cdots, 1.5\}$, $\beta_{d}$ and $\beta_{s}$ in $\{0.5, 1, 1.5, 2\}$, and $\gamma$ in $\{0.5, 0.7, 0.9\}$. All the experiments are conducted on an NVIDIA GTX 1080Ti GPU. 

\paragraph{Results.}
For \textsc{Cora}, \textsc{Citeseer}, and \textsc{Pubmed} datasets, we stick to the public splits (20 nodes per class for training, 1000 nodes for validation, and 500 nodes for testing) used in~\cite{cora}. For \textsc{Coauthor CS} and \textsc{Coauthor Physics}, we follow the splits in~\cite{coauthor}, i.e., 20 labeled nodes per class as the training set, 30 nodes per class as the validation set, and the rest as the test set. The overall results are reported in Table~\ref{tab:results_public} and all the results are obtained over 10 different runs. We can observe significant improvement made by NodeMixup in all the graphs and models. In addition, it is worth noting that PASTEL and ReNode tend to have higher computational costs or increased complexity compared to other methods. It can limit their practical application, particularly when dealing with large graphs. The best result w.r.t each backbone model is shown in boldface. OOM stands for ``out-of-memory."

\subsection{Evaluation on the Large-scale Graph}
\paragraph{Settings.}
We test NodeMixup's effectiveness on the challenging large-scale graph \textsc{ogbn-arxiv} to enhance GNN performance. Baseline models include GCN, GraphSAGE, JKNet~\cite{jknet}, and GCNII~\cite{gcnii}. Data splits come from~\cite{ogb}. Due to GPU limitations, we use GCN and GraphSAGE as backbone models, employing identical configurations as outlined in the source code from~\cite{ogb}.

\paragraph{Results.}
The results are provided in Table~\ref{tab:ogb}. Notably, NodeMixup demonstrates its efficacy in bolstering the performance of fundamental GNN models, such as GCN and GraphSAGE. This enhancement remains pronounced even in the context of large-scale graphs, effectively surpassing the performance of more intricate deep GNN models, i,e, GCNII. This observation underscores the ability of NodeMixup to yield substantial gains in predictive accuracy while maintaining efficiency. The baseline performance are obtained from the OGB Leaderboards (\url{https://ogb.stanford.edu/docs/leader_nodeprop/}).

\begin{table}[!th]
    \centering
    \begin{tabular}{l c c c}
    \hline
    \toprule
          \multirow{2}{*}{Models} & \multicolumn{2}{c}{Accuracy (\%)} \\
          ~ & Test & Valid &\\
          \midrule
          GCN & 71.74\small{$\pm$0.2} & 73.00\small{$\pm$0.1} \\
          GraphSAGE & 71.49\small{$\pm$0.2} & 72.77\small{$\pm$0.1} \\
          JKNet & 72.19\small{$\pm$0.2} & 73.35\small{$\pm$0.1} \\
          GCNII & 72.74\small{$\pm$0.1} & - \\
          \midrule
          GCN+NodeMixup & \textbf{73.46$\pm$\small{0.2}} & 74.13$\pm$\small{0.2} \\
		  GraphSAGE+NodeMixup & 73.24$\pm$\small{0.2} & \textbf{74.14$\pm$\small{0.1}} \\
    \bottomrule
    \hline
    \end{tabular}
    \caption{Node Classification on \textsc{ogbn-arxiv}.}
    \label{tab:ogb}
\end{table}

\subsection{Ablation Analysis}
\label{sec:abla}
NodeMixup comprises three key modules: cross-set mixup, class-specific mixup, and NLD-aware sampling. In Table~\ref{tab:ablation}, we conduct an investigation into each design utilizing the \textsc{Cora} dataset and GCN as the underlying model. Note that, except for the design under examination, NodeMixup remains unchanged. Our observations reveal several fascinating properties, which are outlined below.

 \paragraph{Cross-set Pairing.} The traditional mixup~\cite{mixup} approach samples mixed pairs exclusively from the labeled set, making it function primarily as a data augmentation technique without addressing the under-reaching problem. Consequently, it does not generate significant improvements since a considerable number of unlabeled nodes remain inaccessible. In contrast, our labeled-unlabeled (LU) pairing, as opposed to labeled-labeled (LL) and labeled-all (LA) pairings, allows GNNs to leverage information from a larger pool of unlabeled nodes. This facilitates distribution alignment and ultimately leads to enhanced performance.
   
 \paragraph{Class-specific Mixup.} Table~\ref{tab:ablation} reveals that both inter-class (IE) and intra-class (IR) interpolations contribute to performance enhancements when applied between nodes. Nonetheless, solely interpolating nodes with the same label does not significantly improve the backbone model, aligning with findings in~\cite{mixup}. It is because such an approach fails to take advantage of the mutually beneficial neighbor information among intra-class nodes.
 
 \paragraph{NLD-aware Sampling.} To investigate the low-degree biased sampling strategy employed in NodeMixup, we conducted a comparison with random sampling. Our findings consistently demonstrated that our sampling strategy consistently outperforms random sampling. This can be attributed to two factors: (1) NLD similarity guarantees a higher probability of mixing nodes with similar neighbor patterns, thereby enhancing information alignment and generalization; (2) assigning larger sampling weights to low-degree nodes, which are typically difficult to access in a general scenario.
   
\begin{table}[ht!]
    \centering
    \begin{tabular}{ c c c}
    \hline
    \toprule
    \multicolumn{2}{c}{GCN} & 81.57\small{$\pm$0.4} \\
	\multicolumn{2}{c}{GCN + vanilla mixup} & 81.67\small{$\pm$0.3} \\
	
	\midrule
	\midrule
	
\multirow{3}{*}{Cross-set Pairing} & LL pairing & 81.85\small{$\pm$0.2} \\
 ~ & LA pairing & 81.97\small{$\pm$0.6} \\
 ~ & \textbf{LU pairing} & \textbf{83.64\small{$\pm$0.5}} \\
 
 \midrule
 
 \multirow{3}{*}{Class-specific Mixup} & IE mixup & 81.97\small{$\pm$0.3} \\
 ~ & IR mixup & 82.06\small{$\pm$0.3} \\
 ~ & \textbf{IE-\&IR-class mixup} & \textbf{83.64\small{$\pm$0.5}} \\
 
 \midrule
 
 \multirow{2}{*}{NLD-aware Sampling} & Random & 82.41\small{$\pm$0.2} \\
 ~ & \textbf{NLD-aware} & \textbf{83.64\small{$\pm$0.5}} \\
	
    \bottomrule
    \hline
    \end{tabular}
    \caption{Ablation Analysis.}
    \label{tab:ablation}
\end{table}

\section{Conclusion}
In this work, we take an in-depth look at the under-reaching issue through comprehensive empirical investigations and extensive experimental analysis on widely-used graphs. The issue widely exists in various GNN models and degrades their performance. Our findings shed light on the fact that the reachability of GNNs should be further strengthened. To this purpose, we propose an architecture-agnostic framework, NodeMixup, to improve reachability for GNNs. It effectively addresses the under-reaching issue, helping GNNs to achieve better performance using different graphs.

\section{Acknowledgments}
This work was supported in part by the National Natural Science Foundation of China under Grants 62133012, 61936006, 62073255, and 62303366, in part by the Innovation Capability Support Program of Shaanxi under Grant 2021TD-05, and in part by the Key Research and Development Program of Shaanxi under Grant 2020ZDLGY04-07.

\bibliography{aaai24}

\begin{thebibliography}{43}
\providecommand{\natexlab}[1]{#1}

\bibitem[{Alon and Yahav(2020)}]{bottleneck}
Alon, U.; and Yahav, E. 2020.
\newblock On the bottleneck of graph neural networks and its practical implications.
\newblock \emph{arXiv preprint arXiv:2006.05205}.

\bibitem[{Barcel{\'o} et~al.(2020)Barcel{\'o}, Kostylev, Monet, P{\'e}rez, Reutter, and Silva}]{under_reaching}
Barcel{\'o}, P.; Kostylev, E.~V.; Monet, M.; P{\'e}rez, J.; Reutter, J.; and Silva, J.-P. 2020.
\newblock The logical expressiveness of graph neural networks.
\newblock In \emph{8th International Conference on Learning Representations (ICLR 2020)}.

\bibitem[{Berthelot et~al.(2019)Berthelot, Carlini, Goodfellow, Papernot, Oliver, and Raffel}]{sharp}
Berthelot, D.; Carlini, N.; Goodfellow, I.; Papernot, N.; Oliver, A.; and Raffel, C.~A. 2019.
\newblock Mixmatch: A holistic approach to semi-supervised learning.
\newblock \emph{Advances in neural information processing systems}, 32.

\bibitem[{Bishop and Nasrabadi(2006)}]{cross_entropy}
Bishop, C.~M.; and Nasrabadi, N.~M. 2006.
\newblock \emph{Pattern recognition and machine learning}, volume~4.
\newblock Springer.

\bibitem[{Br{\"u}el-Gabrielsson, Yurochkin, and Solomon(2022)}]{rewiring}
Br{\"u}el-Gabrielsson, R.; Yurochkin, M.; and Solomon, J. 2022.
\newblock Rewiring with positional encodings for graph neural networks.
\newblock \emph{arXiv preprint arXiv:2201.12674}.

\bibitem[{Buchnik and Cohen(2018)}]{influence_decay}
Buchnik, E.; and Cohen, E. 2018.
\newblock Bootstrapped graph diffusions: Exposing the power of nonlinearity.
\newblock In \emph{Abstracts of the 2018 ACM International Conference on Measurement and Modeling of Computer Systems}, 8--10.

\bibitem[{Chen et~al.(2021)Chen, Lin, Zhao, Ren, Li, Zhou, and Sun}]{renode}
Chen, D.; Lin, Y.; Zhao, G.; Ren, X.; Li, P.; Zhou, J.; and Sun, X. 2021.
\newblock Topology-imbalance learning for semi-supervised node classification.
\newblock \emph{Advances in Neural Information Processing Systems}, 34: 29885--29897.

\bibitem[{Chen et~al.(2020)Chen, Wei, Huang, Ding, and Li}]{gcnii}
Chen, M.; Wei, Z.; Huang, Z.; Ding, B.; and Li, Y. 2020.
\newblock Simple and deep graph convolutional networks.
\newblock In \emph{International Conference on Machine Learning}, 1725--1735. PMLR.

\bibitem[{Chien et~al.(2021)Chien, Peng, Li, and Milenkovic}]{gprgnn}
Chien, E.; Peng, J.; Li, P.; and Milenkovic, O. 2021.
\newblock Adaptive Universal Generalized PageRank Graph Neural Network.
\newblock \emph{ICLR}.

\bibitem[{Crisostomi et~al.(2022)Crisostomi, Antonelli, Maiorca, Moschella, Marin, and Rodol{\`a}}]{mixup_metric_g}
Crisostomi, D.; Antonelli, S.; Maiorca, V.; Moschella, L.; Marin, R.; and Rodol{\`a}, E. 2022.
\newblock Metric Based Few-Shot Graph Classification.
\newblock \emph{arXiv preprint arXiv:2206.03695}.

\bibitem[{Defferrard, Bresson, and Vandergheynst(2016)}]{chebnet}
Defferrard, M.; Bresson, X.; and Vandergheynst, P. 2016.
\newblock Convolutional neural networks on graphs with fast localized spectral filtering.
\newblock \emph{Advances in neural information processing systems}, 29.

\bibitem[{Di~Giovanni et~al.(2023)Di~Giovanni, Giusti, Barbero, Luise, Lio, and Bronstein}]{over}
Di~Giovanni, F.; Giusti, L.; Barbero, F.; Luise, G.; Lio, P.; and Bronstein, M. 2023.
\newblock On Over-Squashing in Message Passing Neural Networks: The Impact of Width, Depth, and Topology.
\newblock \emph{arXiv preprint arXiv:2302.02941}.

\bibitem[{Fey and Lenssen(2019)}]{torch_geo}
Fey, M.; and Lenssen, J.~E. 2019.
\newblock Fast Graph Representation Learning with {PyTorch Geometric}.
\newblock In \emph{ICLR Workshop on Representation Learning on Graphs and Manifolds}.

\bibitem[{Gilmer et~al.(2017)Gilmer, Schoenholz, Riley, Vinyals, and Dahl}]{mpnn}
Gilmer, J.; Schoenholz, S.~S.; Riley, P.~F.; Vinyals, O.; and Dahl, G.~E. 2017.
\newblock Neural message passing for quantum chemistry.
\newblock In \emph{ICML}, 1263--1272. PMLR.

\bibitem[{Guo and Mao(2021)}]{ifmixup_g}
Guo, H.; and Mao, Y. 2021.
\newblock ifmixup: Towards intrusion-free graph mixup for graph classification.
\newblock \emph{arXiv e-prints}, arXiv--2110.

\bibitem[{Hamilton, Ying, and Leskovec(2017)}]{sage}
Hamilton, W.; Ying, Z.; and Leskovec, J. 2017.
\newblock Inductive representation learning on large graphs.
\newblock \emph{Advances in neural information processing systems}, 30.

\bibitem[{Hu et~al.(2020)Hu, Fey, Zitnik, Dong, Ren, Liu, Catasta, and Leskovec}]{ogb}
Hu, W.; Fey, M.; Zitnik, M.; Dong, Y.; Ren, H.; Liu, B.; Catasta, M.; and Leskovec, J. 2020.
\newblock Open graph benchmark: Datasets for machine learning on graphs.
\newblock \emph{Advances in neural information processing systems}, 33: 22118--22133.

\bibitem[{Kingma and Ba(2015)}]{adam}
Kingma, D.; and Ba, J. 2015.
\newblock Adam: A method for stochastic optimization.
\newblock In \emph{ICLR}.

\bibitem[{Kipf and Welling(2017)}]{gcn}
Kipf, N.~T.; and Welling, M. 2017.
\newblock Semi-Supervised Classification with Graph Convolutional Networks.
\newblock \emph{international conference on learning representations}.

\bibitem[{Klicpera, Bojchevski, and Günnemann(2019)}]{appnp}
Klicpera, J.; Bojchevski, A.; and Günnemann, S. 2019.
\newblock Predict then Propagate: Graph Neural Networks meet Personalized PageRank.
\newblock \emph{ICLR}.

\bibitem[{Kornblith et~al.(2019)Kornblith, Norouzi, Lee, and Hinton}]{cka}
Kornblith, S.; Norouzi, M.; Lee, H.; and Hinton, G. 2019.
\newblock Similarity of neural network representations revisited.
\newblock In \emph{International Conference on Machine Learning}, 3519--3529. PMLR.

\bibitem[{Lee et~al.(2013)}]{pl}
Lee, D.-H.; et~al. 2013.
\newblock Pseudo-label: The simple and efficient semi-supervised learning method for deep neural networks.
\newblock In \emph{Workshop on challenges in representation learning, ICML}, volume~3, 896.

\bibitem[{Li, Han, and Wu(2018)}]{li2018deeper}
Li, Q.; Han, Z.; and Wu, X.-M. 2018.
\newblock Deeper insights into graph convolutional networks for semi-supervised learning.
\newblock In \emph{Thirty-Second AAAI conference on artificial intelligence}.

\bibitem[{Lu et~al.(2021)Lu, Zhan, Guan, Liu, Yu, Zhao, Yang, and Tao}]{skipnode}
Lu, W.; Zhan, Y.; Guan, Z.; Liu, L.; Yu, B.; Zhao, W.; Yang, Y.; and Tao, D. 2021.
\newblock SkipNode: On Alleviating Over-smoothing for Deep Graph Convolutional Networks.
\newblock \emph{arXiv preprint arXiv:2112.11628}.

\bibitem[{Navarro and Segarra(2022)}]{graphmad_g}
Navarro, M.; and Segarra, S. 2022.
\newblock GraphMAD: Graph Mixup for Data Augmentation using Data-Driven Convex Clustering.
\newblock \emph{arXiv preprint arXiv:2210.15721}.

\bibitem[{Oono and Suzuki(2020)}]{oono2020graph}
Oono, K.; and Suzuki, T. 2020.
\newblock Graph Neural Networks Exponentially Lose Expressive Power for Node Classification.
\newblock \emph{ICLR}.

\bibitem[{Park, Shim, and Yang(2022)}]{mixup_transplant_g}
Park, J.; Shim, H.; and Yang, E. 2022.
\newblock Graph transplant: Node saliency-guided graph mixup with local structure preservation.
\newblock In \emph{Proceedings of the AAAI Conference on Artificial Intelligence}, volume~36, 7966--7974.

\bibitem[{Rizve et~al.(2021)Rizve, Duarte, Rawat, and Shah}]{ups}
Rizve, M.~N.; Duarte, K.; Rawat, Y.~S.; and Shah, M. 2021.
\newblock In defense of pseudo-labeling: An uncertainty-aware pseudo-label selection framework for semi-supervised learning.
\newblock \emph{arXiv preprint arXiv:2101.06329}.

\bibitem[{Shchur et~al.(2018)Shchur, Mumme, Bojchevski, and G{\"u}nnemann}]{coauthor}
Shchur, O.; Mumme, M.; Bojchevski, A.; and G{\"u}nnemann, S. 2018.
\newblock Pitfalls of graph neural network evaluation.
\newblock \emph{arXiv preprint arXiv:1811.05868}.

\bibitem[{Shimodaira(2000)}]{distribution_shift}
Shimodaira, H. 2000.
\newblock Improving predictive inference under covariate shift by weighting the log-likelihood function.
\newblock \emph{Journal of statistical planning and inference}, 90(2): 227--244.

\bibitem[{Sun et~al.(2022)Sun, Li, Yuan, Fu, Peng, Ji, Li, and Yu}]{pastel}
Sun, Q.; Li, J.; Yuan, H.; Fu, X.; Peng, H.; Ji, C.; Li, Q.; and Yu, P.~S. 2022.
\newblock Position-aware structure learning for graph topology-imbalance by relieving under-reaching and over-squashing.
\newblock In \emph{Proceedings of the 31st ACM International Conference on Information \& Knowledge Management}, 1848--1857.

\bibitem[{Topping et~al.(2021)Topping, Giovanni, Chamberlain, Dong, and Bronstein}]{understanding-os}
Topping, J.; Giovanni, D.~F.; Chamberlain, P.~B.; Dong, X.; and Bronstein, M.~M. 2021.
\newblock Understanding over-squashing and bottlenecks on graphs via curvature.
\newblock \emph{ICLR 2022}.

\bibitem[{Velickovic et~al.(2018)Velickovic, Cucurull, Casanova, Romero, Liò, and Bengio}]{gat}
Velickovic, P.; Cucurull, G.; Casanova, A.; Romero, A.; Liò, P.; and Bengio, Y. 2018.
\newblock Graph Attention Networks.
\newblock \emph{ICLR}.

\bibitem[{Verma et~al.(2021)Verma, Qu, Kawaguchi, Lamb, Bengio, Kannala, and Tang}]{graphmix}
Verma, V.; Qu, M.; Kawaguchi, K.; Lamb, A.; Bengio, Y.; Kannala, J.; and Tang, J. 2021.
\newblock Graphmix: Improved training of gnns for semi-supervised learning.
\newblock In \emph{Proceedings of the AAAI conference on artificial intelligence}, volume~35, 10024--10032.

\bibitem[{Wang et~al.(2021)Wang, Wang, Liang, Cai, and Hooi}]{mixup_for_node}
Wang, Y.; Wang, W.; Liang, Y.; Cai, Y.; and Hooi, B. 2021.
\newblock Mixup for node and graph classification.
\newblock In \emph{Proceedings of the Web Conference 2021}, 3663--3674.

\bibitem[{Wu et~al.(2019)Wu, Souza, Zhang, Fifty, Yu, and Weinberger}]{sgc}
Wu, F.; Souza, A.; Zhang, T.; Fifty, C.; Yu, T.; and Weinberger, K. 2019.
\newblock Simplifying graph convolutional networks.
\newblock In \emph{International conference on machine learning}, 6861--6871. PMLR.

\bibitem[{Wu et~al.(2021)Wu, Lin, Gao, Tan, Li et~al.}]{graphmixup}
Wu, L.; Lin, H.; Gao, Z.; Tan, C.; Li, S.; et~al. 2021.
\newblock Graphmixup: Improving class-imbalanced node classification on graphs by self-supervised context prediction.
\newblock \emph{arXiv preprint arXiv:2106.11133}.

\bibitem[{Xu et~al.(2019)Xu, Hu, Leskovec, and Jegelka}]{gin}
Xu, K.; Hu, W.; Leskovec, J.; and Jegelka, S. 2019.
\newblock How Powerful are Graph Neural Networks?
\newblock \emph{international conference on learning representations}.

\bibitem[{Xu et~al.(2018)Xu, Li, Tian, Sonobe, Kawarabayashi, and Jegelka}]{jknet}
Xu, K.; Li, C.; Tian, Y.; Sonobe, T.; Kawarabayashi, K.-i.; and Jegelka, S. 2018.
\newblock Representation learning on graphs with jumping knowledge networks.
\newblock In \emph{International Conference on Machine Learning}, 5453--5462. PMLR.

\bibitem[{Yang, Cohen, and Salakhudinov(2016)}]{cora}
Yang, Z.; Cohen, W.; and Salakhudinov, R. 2016.
\newblock Revisiting semi-supervised learning with graph embeddings.
\newblock In \emph{International conference on machine learning}, 40--48. PMLR.

\bibitem[{Zhang et~al.(2017)Zhang, Cisse, Dauphin, and Lopez-Paz}]{mixup}
Zhang, H.; Cisse, M.; Dauphin, Y.~N.; and Lopez-Paz, D. 2017.
\newblock mixup: Beyond empirical risk minimization.
\newblock \emph{arXiv preprint arXiv:1710.09412}.

\bibitem[{Zheng et~al.(2022)Zheng, Xia, Zhang, Kharlamov, and Dong}]{distribution}
Zheng, Q.; Xia, X.; Zhang, K.; Kharlamov, E.; and Dong, Y. 2022.
\newblock On the distribution alignment of propagation in graph neural networks.
\newblock \emph{AI Open}, 3: 218--228.

\bibitem[{Zhu et~al.()Zhu, Yan, Heimann, Zhao, Akoglu, and Koutra}]{NLD}
Zhu, J.; Yan, Y.; Heimann, M.; Zhao, L.; Akoglu, L.; and Koutra, D. ????
\newblock Heterophily and Graph Neural Networks: Past, Present and Future.
\newblock \emph{Data Engineering}, 10.

\end{thebibliography}

\end{document}